\documentclass[runningheads]{llncs}
\usepackage{graphicx}
\usepackage{amsmath,amssymb} 
\usepackage{color}
\usepackage{multirow}
\usepackage{graphicx}
\usepackage{hyperref}
\hypersetup{
    colorlinks=true,
    linkcolor=blue,
    filecolor=magenta,      
    urlcolor=cyan,
}

\begin{document}
\pagestyle{headings}
\mainmatter



\title{3D Human Motion Estimation via \\Motion Compression and Refinement} 


\titlerunning{MEVA}
%
\author{Zhengyi Luo \orcidID{0000-0002-1842-7622} \and S. Alireza Golestaneh \orcidID{0000-0002-3230-2797} \and Kris M. Kitani \orcidID{0000-0002-9389-4060}}
\authorrunning{Z. Luo et al.}
%
\institute{Carnegie Mellon University, Pittsburgh, PA 15213, USA  \\
\email{\{zluo2,sgolesta,kkitani\}@cs.cmu.edu}}

\maketitle

\begin{figure}[ht]
    \centering
    \includegraphics [scale=.47]{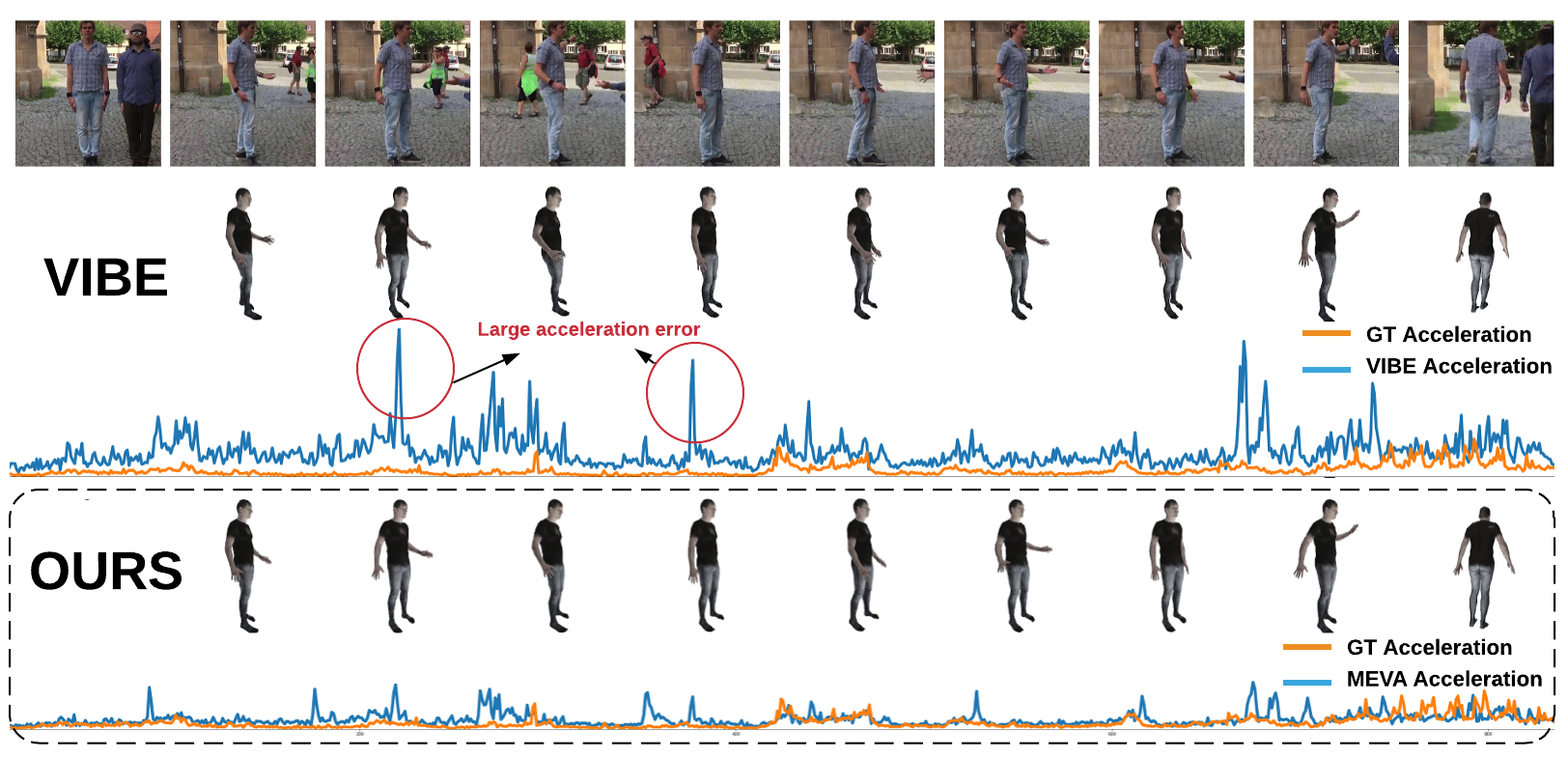}
    \caption{Given an in-the-wild video, state-of-the-art methods   (\textit{e.g.,} VIBE \cite{vibe})  can achieve high per joint accuracy, but also suffers from high acceleration error. To tackle this, we develop a two-stage motion estimation method, \textit{MEVA}, that is able to produce both accurate and smooth human motion.}
    \label{fig:teaser}
\end{figure}

\begin{abstract}
We develop a technique for generating smooth and accurate 3D human pose and motion estimates from RGB video sequences. Our method, which we call \textbf{M}otion \textbf{E}stimation via \textbf{V}ariational \textbf{A}utoencoder (\textit{MEVA}), decomposes a temporal sequence of human motion into a smooth motion representation using auto-encoder-based motion compression and a residual representation learned through motion refinement. This two-step encoding of human motion captures human motion in two stages: a general human motion estimation step that captures the coarse overall motion, and a residual estimation that adds back person-specific motion details. Experiments show that our method produces both smooth and accurate 3D human pose and motion estimates.
\end{abstract}


\section{Introduction}
\label{Introduction}


Estimating the 3D pose sequence of a person from a single video requires a computational model that can extract the underlying kinematics of human motion while also preserving motions that are unique to the person being captured. Since people share a similar body structure (\emph{e.g.,} same number of joints) and similar physical constraints (\emph{e.g.,} joint limitations), it is possible to learn a generalized kinematic model that can be matched against the image to infer the general motion of a person. However, since generalized models of motion can also fail to model person-specific motions, it may also be necessary to `add back in' or refine the general motion estimates using image evidence. In this work, we present a two-stage 3D motion estimation method that first extracts coarse kinematic sequences of a person in a video and then refines that sequence to produce a more accurate 3D motion estimate. We show that by decomposing the inference process into (1) a general model of motion and (2) a person-specific model of motion, we are able to obtain more accurate and smooth estimates.

Over the past years, significant progress has been made on improving the accuracy of 3D human pose estimation. Impressive results have been obtained through human mesh recovery from single images  \cite{1908.07172,spin,Georgakis,DensePose,DenseRaC} and videos \cite{hmmr,vibe,hmr,Hossain2018}. The main metric used to evaluate these methods is the Mean Per Joint Position Error (MPJPE), which measures the performance in terms of the relative joint positions computed for each frame of a video. However, less emphasis has been given to the temporal smoothness of the estimated motion. Optimizing for this metric, the tendency is to generate pose estimates that `jitter' near the true pose. This is expected as the MPJPE only penalizes for spatial errors and is not designed to account for temporal consistency. As the methods for 3D pose estimation have improved in recent years, the `jitter' has become less pronounced, especially when applied to dynamic scenes with vibrant moving backgrounds and camera motion. However, by rendering the estimated 3D pose of state-of-the-art methods on a plain background, the `jitter' can still be observed, resulting in an overall unnatural motion estimation. 

The issue of temporal smoothness is a known problem and has been addressed in part by prior work. Large-scale motion datasets such as Archive of Motion Capture as Surface Shapes (AMAAS) \cite{amass}) and adversarial loss have enabled methods to improve both pose accuracy and temporal smoothness \cite{vibe,hmmr}. Other methods have been developed to enforce temporal smoothness \cite{1908.07172} by letting the model predict frame ordering. However, using prior knowledge only in the loss function, it is hard to find the balance between smoothness and accuracy.

In this work, we argue that striking such balance between smoothness and accuracy can be done through an explicit breakdown of coarse and fine motion. First, we learn a coarse motion model by observing a large dataset of human motion--since human motion is typically smooth (\emph{e.g.}, we usually do not shake as we walk), if one were to fit a model to a large set of human motions, most of the data would lie in a sub-space in which motions are smooth. This implies that if we were to compress human motion data, it should learn a latent subspace in which the motions are inherently smooth and coarse. Using this latent space as the regression target, we can directly infer coarse human motion from the input videos. The problem of using such a human motion subspace, of course, is that rare motions (\emph{e.g.,} sudden motions) are removed from the motion data. To retain such motion, we also argue that producing the final 3D motion estimate can be treated as a refinement step to "add back" the fine details to the coarse motion sequence. 

To validate our arguments, we propose a two-stage 3D human motion estimation method that first estimates a coarse human pose sequence through data compression using a Variational Autoencoder (VAE), which we call the Variational Motion Estimator (VME). Then we take the output of the VME and refine the pose estimate using the image evidence with a pose regressor, which we call the Motion Refinement Regressor (MRR).

In summary, we propose a video-based 3D human motion estimation method that focuses on producing smooth and accurate human motion sequences. Our main contributions are as follows: (1) we propose a two-stage motion estimation method for ensuring temporal smoothness and accurate pose estimates; (2) we describe a procedure to learn a robust Variational Autoencoder that can serve as a latent human motion subspace for estimating coarse 3D human motion from videos; (3) we demonstrate state-of-the-art pose/motion estimation performance on challenging in-the-wild dataset such as 3DPW \cite{3dpw}, reducing acceleration error by $54.3\%$ while achieving state-of-the-art MPJPE results.

\section{Related Works}
\label{Related}
In this section, we will first review the relevant work in human shape and pose recovery from a single image and from videos--human motion recovery can be treated as a subset of human pose estimation, as human motion is a sequence of human poses. Then we will review how existing methods use human motion as a prior and how popular methods map motion sequences to a low-dimensional space. 

\subsection{Recovering 3D human pose and shape from a single image}  
Here we focus on model-based methods that jointly recover human shape and pose. We choose to use a parametric 3D human body model \cite{smplx,SMPL,SCAPE} since it can be easily turned into a 3D human mesh that is usable for downstream tasks such as animation. Directly fitting a parametric 3D human body to image input has gained substantial traction over the years, morphing from methods that require silhouette or human input \cite{Grauman2003,Agarwal2006,Sigal2009,Zhou2010,5459300}, to ones that can directly fit model parameters to 2D joint positions \cite{smplify}, and to ones that can directly estimate shape and pose from images \cite{hmr,Omran2018,Guler2019,smplx,Tan2017,Tung2017,spin}. Due to the lack of ground truth 3D labels, these methods use a weakly supervised approach to fit the 3D human body to 2D joint positions \cite{hmr,Tan2017,Tung2017}, body part segmentation \cite{Omran2018,Pavlakos2018}, or dense pixel correspondence \cite{DenseRaC}. Although these methods achieve amazing results, their extracted motion tends to be unstable due to the lack of temporal information. 

\subsection{Recovering 3D human pose and shape from video}
Using temporal information to aid 3D human pose estimation is a natural extension to single frame methods. \cite{Hossain2018,Pavllo2019,Dabral2018,Xu} focus on "lifting" predicted joint positions from 2D to 3D, and uses LSTM \cite{Hossain2018}, Temporal Convolution \cite{Pavllo2019}, and fully connected layers  \cite{Dabral2018} to exploit temporal information. \cite{Mehta2017,Mehta2019}, on the other hand, predict 3D joint positions directly from images and use a temporal filter to postprocess the motion sequence. For methods that jointly recover shape and pose. HMMR \cite{hmmr}, Sun \textit{et al.} \cite{1908.07172} and VIBE \cite{vibe} are the best performing models that exploit temporal information. HMMR \cite{hmmr} proposes to enforce temporal consistency by letting the model predict future and past frames of motion. Sun \textit{et al.} \cite{1908.07172} learns temporal information by predicting the ordering of shuffled frames. VIBE \cite{vibe} utilizes temporal information by employing a Gated Recurrent Unit (GRU) to convert the input frames of features into a sequence of temporally correlated latent features. 

\subsection{Human Pose and Motion Prior}
Using prerecorded human motion sequences as a prior has also been explored in various tasks related to human motion. Earlier work like \cite{Ren2005} tries to quantify unnaturalness in animated human motion by statistically analyzing existing motion capture (MoCap) sequences, and \cite{Urtasun2006,Ormoneit2001} propose to use learned MoCap motions to aid 3D motion tracking. More recently, \cite{hmr,hmmr} use an adversarial discriminator at a per-frame level to ensure that the recovered pose is a valid human pose. \cite{smplx} uses a pretrained pose VAE's latent space for a similar purpose. \cite{vibe} proposes to use the discriminator at a temporal level, discriminating against a whole motion sequence. All above methods use a pose or motion prior in an adversarial way, utilizing the prior knowledge in the loss function. 

\subsection{Human Motion Representation}
Compressing human motion into a compact latent representation plays an important role in tasks such as human motion generation \cite{Wang2019,Cai_2018_ECCV,Walker_2017_ICCV}, human motion generation across different modalities \cite{Plappert2018,Ahuja,Yamada2018,Xiaodong2019},  and trajectory forecasting \cite{Fragkiadaki2015,Yan2018,diverse,Butepage_2017_CVPR}. Existing methods leverage different generative models such as VAE \cite{Yan2018,diverse}, generative flow \cite{Prediction}, or generative adversarial networks  \cite{Wang2019} to achieve a compact motion representation in the latent space.

\section{Approach}
\begin{figure}[ht]
    \centering
    \includegraphics[width=0.85\linewidth]{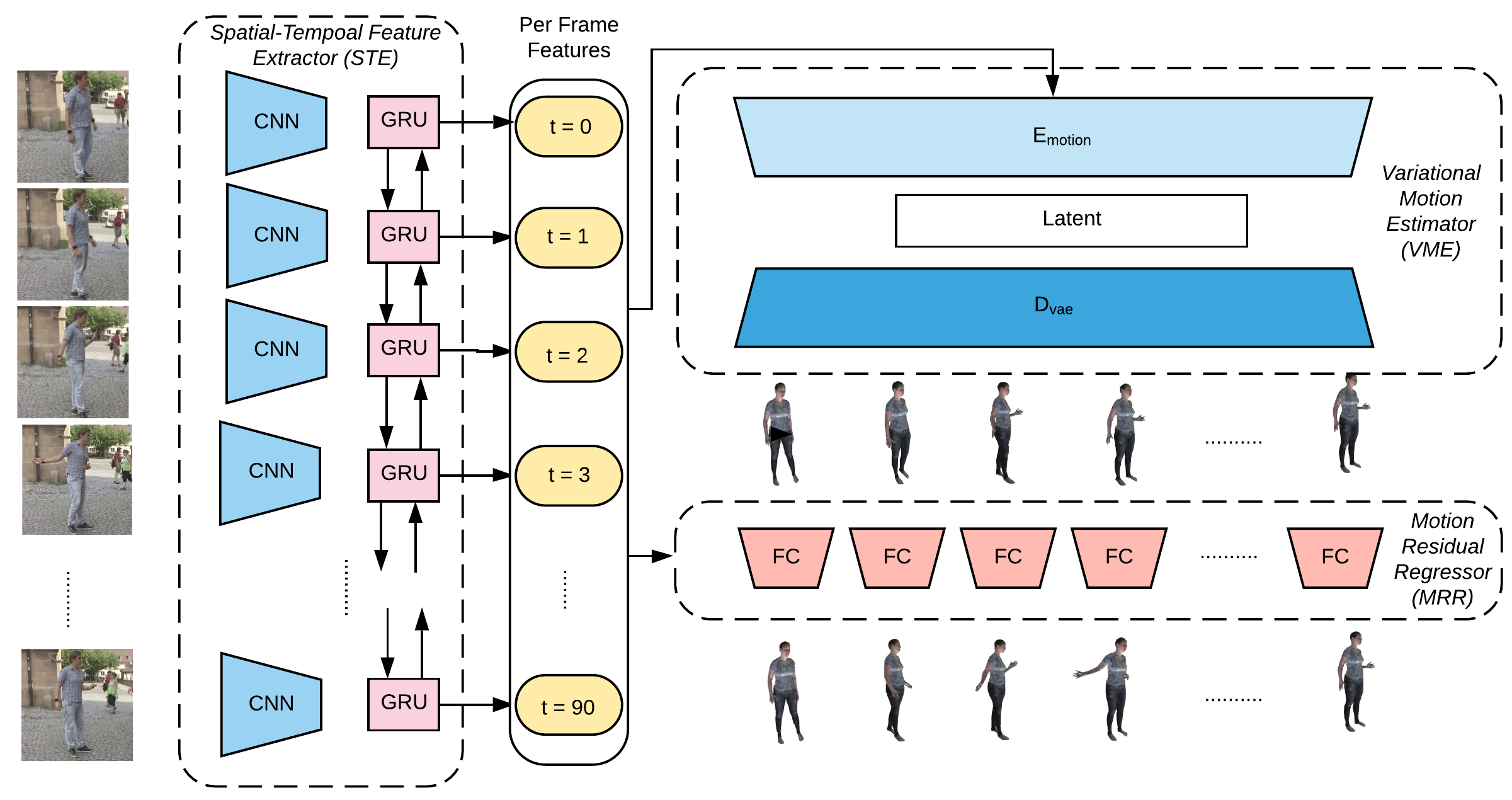}
    \caption{\textbf{Overall Architecture.} \textit{MEVA} estimates motion from videos by first extracting temporal features using Spatio-Temporal Feature Extractor (STE) and then estimates the overall coarse motion inside the video with Variational Motion Estimator (VME). Finally, a Motion Residual Regressor (MRR) is used to refine the motion estimates.}
    \label{fig:archi}
\end{figure}

As discussed in Sections \ref{Introduction} and \ref{Related}, existing human motion estimation methods often find it difficult to achieve a balance between temporal smoothness and accuracy. To tackle this, we propose \textit{MEVA}, \textbf{M}otion \textbf{E}stimation via \textbf{V}ariational \textbf{A}utoencoder, a framework that learns the overall coarse motion first then adds back fine detailed motion as a residual. \textit{MEVA} processes the inputs in three steps: it first extracts correlated temporal features using a spatio-temporal feature extractor (STE), and then captures the overall coarse motion through a variational motion estimator (VME), and finally uses a motion residual regressor (MRR) to add back the fine motion details. Our overall framework can be shown in Fig. \ref{fig:archi}. In this section, we will first setup the overall problem and then present the details of our framework. Finally, we will discuss the training procedures. 

\subsection{Problem Formulation}
Given an input video  $V_T = \{I_t\}_{t = 1}^{T}$, where $I_t$ denotes the $t^{th}$ frame, our task is to recover coherent human motion sequences $M_T = \{\theta_t\}_{t = 1}^{T}$ where each $\theta_t$ represents the human pose for the $t^{th}$ frame. To represent the human motion, we utilize the SMPL 3D human mesh model \cite{SMPL}. We choose SMPL parametrization over 3D joint positions or other human models due to its versatility: SMPL parameters can be easily converted to 3D joint positions and human mesh. Specifically, a human body is represented by its shape $\beta$ and pose $\theta$, denoted by $\Theta = \{\beta, \theta\} $. 
Given $\theta$ and $\beta$, let $S$ denote the pretrained SMPL function, where $S(\Theta): \beta, \theta \rightarrow R^{6890 \times 3}$ ( $6890$ is the number of vertices of the resulting triangular human mesh). The pose parameter $\theta \in R^{24 \times N}$ stands for the joint angles for the 23 joints plus the root orientation. $N$ is the dimension of the chosen rotation representation ($N=3$ for axis/euler angle, $N=4$ for quaternions, $N = 6$ for a 6 degrees-of-freedom  rotation representation \cite{rotrep}). The shape parameters $\beta \in R^{10}$ represent the linear coefficient for the principal component of the parametric human shape space. Given a set of $\beta$ and $\theta$, $S$ can recover the 3D joint positions through a pretrained mesh vertex regressor $P: jp^{3d} = P \{S(\beta, \theta)\} \in R^{N \times 3}$. To project the 3D joint positions back to 2D images, a weak perspective camera $\pi = \{s, t_x, t_y\}$ needs to be estimated: $jp^{2d} = \Pi(P\{S(\beta, \theta)\}) \in R^{N \times 2}$.

Intuitively, recovering human motion from video frames does not require recovering the human shape: one can directly learn a mapping from input video frames to the estimated human motion if sufficient paired ground truth data exist. However, videos paired with ground truth motion annotation (SMPL sequences) require professional capture equipment such as a motion capture (MoCap) rig, which is still rather rare compared to annotated 2D pose datasets. In the absence of 3D data samples, it is critical for our model to learn motion sequences in a semisupervised fashion (\textit{i.e.,} from videos with  3D or 2D labeled joint positions following the approach utilized in \cite{hmr,hmmr,vibe,spin,1908.07172}). 

Overall, our motion estimation objective is to learn a function $\text{MEVA}(V):V_T\rightarrow{M_T}$ where $V_T = \{I_t\}_{t = 1}^{T}$ and   $M_T = \{\theta_t\}_{t = 1}^{T}$.

\subsection{Spatio-Temporal Feature Extractor (STE)}
Human motion is inherently temporal and correlated, and past movement can give cues about future motion. Thus, instead of extracting per-frame visual features using a feed-forward convolutional network independently, we can produce temporally correlated features that lead to better motion sequence modeling. Similar to \cite{vibe}, we use a GRU-based temporal feature extractor (STE) that encodes the input video frames $I_1, I_2, I_3, ... I_T$ into a sequence of temporally correlated features $f'_1, f'_2, f'_3, ... f'_T$. 

\subsection{Variational Motion Estimator (VME)}
\label{compact}
\subsubsection{Human Motion VAE}

To learn a human motion subspace that can encapsulate a broad spectrum of human motion, we choose to use a Variational Autoencoder (VAE). VAEs can effectively capture a large number of possible data modes by explicitly mapping each data point to a latent code, and by imposing a Gaussian prior on the learned latent space, similar motions' latent codes will be near each other. Thus, the VAE's latent space allows for more overlap between codes and therefore enforces smoothness in the latent space. Having a smooth latent code is essential in improving the generalizability of the model since the space of possible human motion is highly correlated and limited. Formally, following the previous work on VAEs  \cite{diverse,Kingma2014,Walker2017,Walker2017}, the objective is to maximize the evidence lower bound of the log-likelihood $p_{\lambda} (x) $ ($\lambda$ and $\phi$ denotes the function parametrization): 

\begin{equation}
    L_{VAE}  = E_{q_{\phi}}[\log p(x|z)] -  \text{KLD}(q_{\phi}(z|x) || p_{\phi}(z)) \label{vae:1},
\end{equation}
where $x$ is the input and the latent code $z \sim \mathcal{N}(\mathbf{0}, \mathbf{I})$.

\begin{figure}[ht]
    \centering
    \includegraphics[scale=0.5]{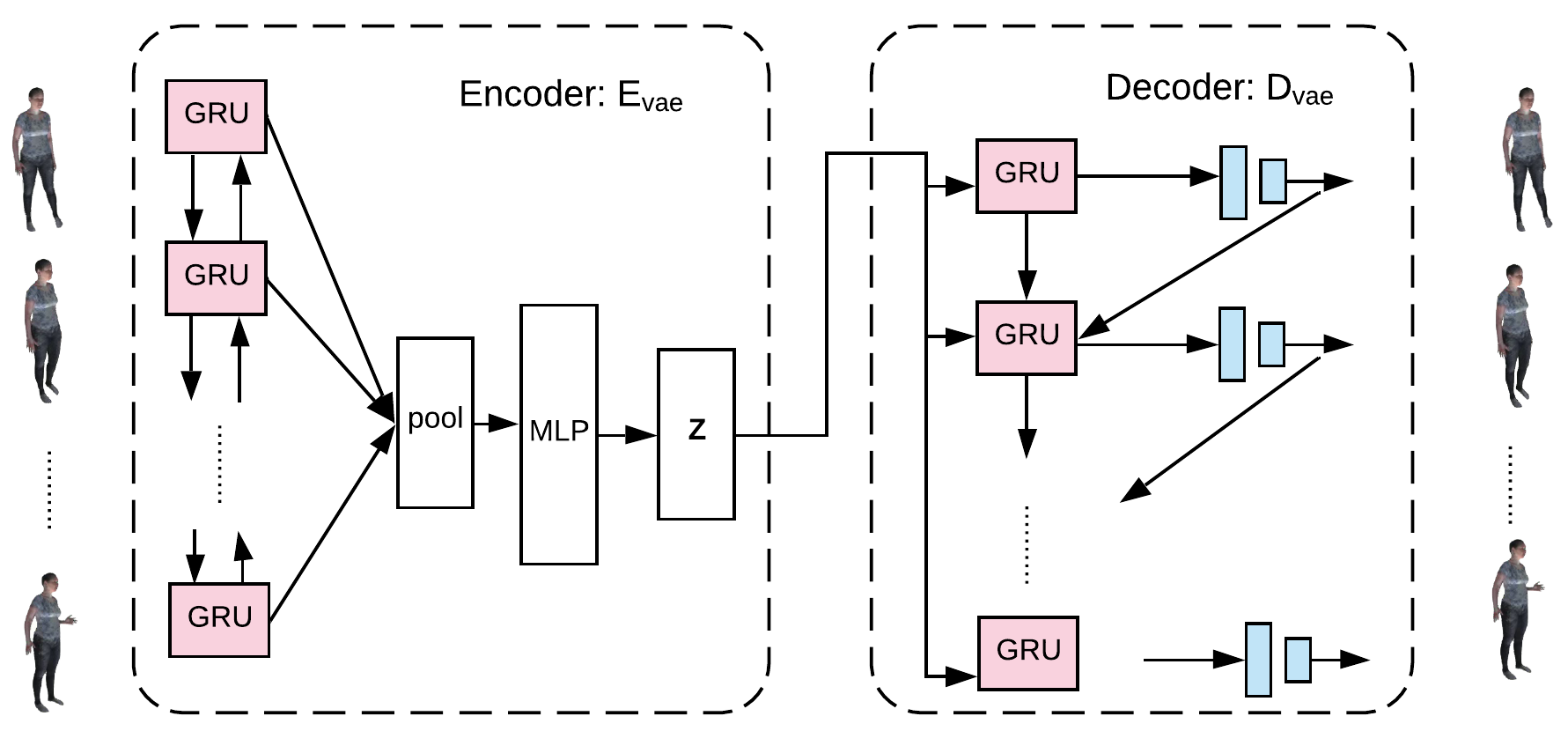}
    \caption{\textbf{Motion VAE Architecture.}}
    \label{fig:vae_archi}
\end{figure}

In the context of encoding human motion via VAE, the encoder $E_{vae}$ takes in a sequence of $W$ frames of human motion represented in terms of SMPL pose parameters: $x = M_W = [\theta_{w1}, \theta_{w2}, \theta_{w3},...] \in R^{W \times 144}$ and outputs the latent code $z$. A single frame of SMPL pose is represented in joint rotations, resulting by a $24 \times 6 = 144$ dimensional input (a 6 degrees-of-freedom rotation representation\cite{rotrep} is used for continuity purpose). The decoder $D_{vae}$ takes in the latent code $z$ and reconstructs the motion $\hat{M}_W$. Both the encoder $E_{vae}$ and decoder $D_{vae}$ are implemented as GRUs, and the detailed architectures are given in Fig.\ref{fig:vae_archi}. Based on the Gaussian parameterization of the VAE, the objective function of Eq.(\ref{vae:1}) can be written Eq.(\ref{vae:obj})

\begin{equation}
\label{vae:obj}
\mathcal{L}_{v a e}(\mathbf{x}; \theta, \phi)=-\frac{1}{S} \sum_{s=1}^{S}\left\|\tilde{\mathbf{x}}-\mathbf{x}\right\|^{2}+\beta \cdot \frac{1}{S_{z}} \sum_{j=1}^{S_{z}}\left(1+2 \log \sigma_{j}-\mu_{j}^{2}-\sigma_{j}^{2}\right),
\end{equation}
where $S$ is the number of samples for the current batch, $S_z$ is the dimension of the current latent variable, and $\beta$ is the weighing parameter. Once the VAE is trained and converged to a desirable reconstruction accuracy, the decoder $D_{vae}$ is frozen for later use. During inference, given a latent code $z \in R^{1\times S_z}$, $D_{vae}$ can decode it back into a sequence of human motion: $M_W \in R^{w \times 144}$. 

\subsubsection{Human Motion Data augmentation}
\label{augmentation} 
 Our learned VAE should be able to generalize to unseen human motion sequences and achieve high reconstruction accuracy to ensure that the learned latent space can indeed serve as a comprehensive human motion subspace. Using an already large-scale human motion dataset AMASS \cite{amass} (13k motion samples with varying length), our trained VAE still has poor generalizability on unseen sequences (for details refer to Sec.\ref{abla_augmentation}). Thus, we devise an elaborate data augmentation scheme that can augment the existing motion and produce viable yet distinct human motion sequences. While data augmentation has been studied extensively in image processing, to the best of our knowledge, few attempts have been done in augmenting a human motion dataset. When used in trajectory forecasting \cite{diverse} and human motion generation \cite{Wang2019}, the generalizability of the VAE latent space has not been discussed extensively since the models only need to generate new motion sequences and do not emphasize on the ability to encode unseen motion sequences.

Given a $T$ frame human motion sequence in SMPL parameters $M_T \in R^{T \times 144}$ with a frame-rate $F_{amass}$, we employ the following data augmentation scheme: 
\begin{itemize}
    \item \textbf{Speeding up and slowing down:} based on $F_{amass}$, we can uniformly up-sample or downsample the frames and produce novel sequences that are still plausible and natural human motion. 
    \item \textbf{Flipping left and right:} The same action, performed using either the left or right hand, will remain a valid human motion. Thus, we can follow the kinematic tree of the SMPL model and mirror the motion across the left and right and generate a new motion sequence.
    \item \textbf{Random root rotation:} We randomly sample a root rotation from a unit sphere to capture different root orientations for possible human motion. Different pose estimators may assume different ground planes and coordinate systems, so SMPL parameters usually come in different root orientations. Sampling random root rotation helps the model cope with different possible coordinate frame choices. 
\end{itemize}

\subsubsection{Learning Smooth Motion from videos}
After learning a comprehensive human motion subspace using the VAE, we learn an additional encoder $E_{motion}$ that can directly extract coarse motion sequences from video features, mapping to the same latent space as $E_{vae}$. Given an input sequence of video features $f_W = \{f_w\}_{w = 1}^{W}$, the encoder $E_{motion}$'s task is to compress the input features into a latent code $z$ that best summarizes the current observation as a coarse human motion sequence. We use the pretrained decoder $D_{vae}$ from the motion VAE to force $E_{motion}$ to sample from our pretrained motion subspace. Constraining the latent space of the $E_{motion}$ to a pretrained human motion subspace provides a strong human motion prior that greatly aids the learning process of $E_{motion}$. Combining $E_{motion}$ and $D_{vae}$, we form our Variational Motion Estimator (VME). 

\subsection{Motion Residual Regressor (MRR)}
As noted in the previous section, the learned motion sequences using the VAE's latent space as the target are inherently smooth and coarse,  capturing the overall motion signature of the current video frames through information compression. To capture the details, we utilize a SMPL regressor from \cite{hmr} that can iteratively refine the estimated poses. The regressor takes in an initial pose and shape estimation $\Theta_t$ and the visual feature $\text{f}_{t}$ for a single frame to calculate its estimation $\Theta_t'$ for $k$ iterations. Notice that though \cite{hmr,hmmr,vibe,spin} all utilize the same regressor, \textit{MEVA} uses it in a fundamentally different way--in \cite{hmr,hmmr,vibe,spin}, the regressor is initialized with mean SMPL pose $\Theta_{mean}$. At a sequence level, a regressor initialized uniformly with the mean pose $\Theta_{mean}$ is trying to capture the overall motion in one pass, while in \textit{MEVA}, the regressor is initialized with the computed poses from VME. Thus, the regressor is only tasked to do small cosmetic changes to the coarse estimation, adding back the fine details of the motion lost during our compression step. Similar to \cite{vibe}, the input visual features ${f}_{t}$ to the regressor are encoded using a temporal visual encoder, so even though each frame's estimation is calculated separately at this stage, the visual feature is already temporally correlated. The VME computes the overall coarse motion from videos by using a general model of motion, and the regressor jointly refines motion and human shape estimates, which amounts to adding back the person-specific motion details at a per frame level. We call this regressor the Motion Residual Regressor (MRR). MRR completes the overall framework of our proposed method, as shown in Fig. \ref{fig:archi}.

\subsection{Training and Losses}
Our framework is trained in two stages.
At first, the motion VAE is pretrained. Then STE, VME, and MRR are trained jointly end-to-end. Using videos with various levels of annotation (2D joint positions, 3D joint positions, SMPL parameters), similar to in \cite{hmmr,vibe,spin}, the networks are trained with losses consisting of $L_{2D}$, $L_{3D}$, $L_{SMPL}$, as long as respective data is available. Specifically: 
\begin{align} 
    L_{meva} &= L_{3D}+L_{2D}+L_{SMPL},\\
L_{3 D} &=\Sigma_{t=1}^{T} || jp^{3d}_{t}-\hat{jp^{3d}_{t}} ||_{2}, \\ 
L_{2 D}     &=\Sigma_{t=1}^{T}||jp^{2d}_{t}-\hat{jp^{2d}_{t}}||_{2}, \\
L_{SMPL} &= ||\beta-\hat{\beta}||_{2}+\Sigma_{t=1}^{T}||\theta_{t}-\hat{\theta}_{t}||_{2}.
\end{align}

For implementation details, please refer to the supplementary material. 

\section{Experiments}
To demonstrate the effectiveness of our proposed method, we evaluate our method in terms of  the  overall  accuracy  and  smoothness  of  the  estimated motion  on MPI-INF-3DPH \cite{mpi3d}, 3DPW \cite{3dpw}, and human 3.6M \cite{h36m}.
In the following sections, we will first describe the main datasets used to train and evaluate \textit{MEVA}. Then in Section \ref{Eval} we provide extensive evaluation results. Finally, in Section \ref{abla}, we provide the ablation studies for our proposed method.

\subsection{Datasets}

For the motion VAE, we use motion sequences from AMASS \cite{amass} for training and sequences from 3DPW \cite{3dpw} for evaluation. For training with videos, in addition to the train split of MPI-INF-3DPH \cite{mpi3d}, 3DPW \cite{3dpw}, and human 3.6M \cite{h36m}, which have 3D joint annotation, we also use InstaVariety \cite{hmmr} and PennAction \cite{pennaction} which contain 2D joint annotation.


For training our motion VAE, we use \textbf{AMASS} \cite{amass}. It is a recent dataset that contains a large sample of human motion sequences in SMPL parameters. These sequences are fitted from MoCap sequences using Mosh++ \cite{amass}. There are in total 13k motion sequences with varying length. We use this dataset only for training our motion VAE. 

For training with videos, \textbf{3DPW}  is the only dataset that contains paired SMPL parameters and video sequences (which provide direct supervision to \textit{MEVA}). The videos from this dataset are mostly outdoors and in-the-wild. It uses paired IMU sensors and video input to compute the near ground truth SMPL pose and shape parameters. This is a relatively small dataset and we use the official split in \cite{3dpw} for train, validation, and test. There are in total 60 videos with varying length (24 train, 24 test, 12 val).  \textbf{MPI-INF-3DHP} is a dataset that contains 3D joint position annotation. It is captured using a multiview camera setup, and the 3D joint annotation is calculated through multiview methods \cite{mpi3d}. There are 8 subjects and 16 videos per subject, in total 128 videos with varying length. We use the official test and train split.  \textbf{H3.6M}  is a popular pose estimation dataset that contains 3D joint position annotations, captured indoors with MoCap markers. Notice that a number of previous works \cite{hmr,hmmr,spin,vibe,1908.07172} had access to a near-ground truth SMPL pose and shape parameters calculated by the Mosh \cite{mosh} method. However, this annotation has since been removed from public access due to legal issues. SMPL parameters provide the best supervision, as noted in \cite{examplar}, so for a fair comparison we retrain some of the state-of-the-art methods without such supervision. There are 840 videos in total across 7 subjects in H3.6M and we use the official train/test subject split ([S1, S5, S6, S7, S8] vs [S9, S11]). During preprocessing, we subsample every 5 frames from the dataset. The \textbf{PennAction} dataset contains human annotated ground truth 2D keypoints paired with video sequences. There are in total 2326 videos with varying length. 
\textbf{InstaVariety} dataset contains human annotated pseudo ground truth 2D keypoints paired with video sequences. The 2D keypionts are estimated using openpose\cite{openpose}. There are in total 28,272 videos with varying length. 

\subsection{Evaluation Results and Analysis}
\label{Eval}
\subsubsection{Metrics}
To best capture human motion, we utilize three popular metrics that measure the overall accuracy and smoothness of the motion. Mean per joint position error (MPJPE) and MPJPE after Procrustes Alignment (PA-MPJPE) measure the 3D joint positional discrepancy between the predicted and ground truth 3D joint positions in millimeters ($m m$), and are calculated after aligning the root position (human pelvis) of the 3D joint positions. Both MPJPE and PA-MPJPE serve as the accuracy indicator of the motion estimator. Acceleration error (ACC-ERR), proposed in \cite{hmmr}, measures the difference between the predicted and ground truth 3D acceleration for each keypoint in $m m / s^{2}$. ACC-ERR serves as the major smoothness indicator for the estimated motion sequences. Acceleration is calculated using the finite difference between individual frames. It is imperative to view these metrics jointly: a low position error indicates overall correctness in motion capture and a better acceleration error marks a smooth and natural estimation of human motion.

\subsubsection{Generalization of Motion VAE}

In this section, we study the genealizability of our learned motion VAE. We report the reconstruction error of the VAE on unseen motion sequences from the 3DPW dataset. Table \ref{t:vae_result} shows VAE motion reconstruction error over the different splits of 3DPW. The Motion VAE model is the best performing model that is trained with all three forms of data augmentation techniques. Detailed analysis about the effects of data augmentation can be found in \ref{abla_augmentation}. The result shows that our VAE generalizes well to unseen sequences and the learned subspace can represent the human motion space with reasonable quality.

\begin{table}[!thb]
\caption{\textbf{VAE Reconstruction Test Error on 3DPW dataset} Here, the VAE is tasked to encode and decode unseen motion sequences from the 3DPW dataset, and we calculate our metrics between the ground truth and reconstructed sequences. Motion sequences from different splits of 3DPW have varying difficulties, but are all unseen by our VAE.  }
\label{t:vae_result} 
\centering
\resizebox{\textwidth}{!}{%
\begin{tabular}{l|c|c|c|c|c|c|c|c|c|c|c|c|c|c|r}
\hline
\multicolumn{1}{c|}{\multirow{2}{*}{}} & \multicolumn{3}{c|}{3DPW Train Split } & \multicolumn{3}{c|}{3DPW Val Split } & \multicolumn{3}{c|}{3DPW Test Split} \\ \cline{2-10} 
\multicolumn{1}{c|}{} & MPJPE $\downarrow$ & PA-MPJPE $\downarrow$ &  ACC-ERR $\downarrow$ & MPJPE $\downarrow$ & PA-MPJPE $\downarrow$ &  ACC-ERR $\downarrow$ & MPJPE $\downarrow$ & PA-MPJPE $\downarrow$ &  ACC-ERR $\downarrow$ \\ \hline
     Motion VAE & 72.7 & 52.3 &  8.9 & 72.1  & 52.9 & 9.3 & 58.7 & 43.9 & 8.2 \\
     \hline
\end{tabular}}
\end{table}

\subsubsection{Quantitative Results}

The result in Table \ref{t:motion_result} shows that our method obtains state-of-the-art results on video motion estimation across all three test datasets. 
Overall, our method achieves comparable results in terms of position error (MPJPE and PA-MPJPE) while significantly improving the smoothness (acceleration error), signifying a smoother and more natural motion estimation without sacrificing accuracy. Notice that all methods in italics have access to SMPL parameter annotation to the H3.6M dataset, which has since been removed from the web due to legal reasons. The SMPL parameters provide the most direct supervision for the task, so the performance gain is significant especially on the H3.6M test set. For a more direct comparison, we retrain the previous state-of-the-art method, VIBE \cite{vibe}, using the official implementation with the exact same datasets as ours. On 3DPW, under the same training condition, \textit{MEVA} outperforms VIBE on almost all three metrics while reducing the acceleration error by 54.3\%, 59.3\%, and 41.3 \%, respectively. Even compared to VIBE trained with extra data, our model achieves comparable results in accuracy while sporting a great reduction in acceleration error, except for the H3.6M dataset. Note that the H3.6M dataset contains mainly indoor scenes with limited background variation and models trained with direct SMPL supervision  tend to perform well on this dataset. Compared to HMMR \cite{hmmr}, which is the state-of-the-art on smoothness, our  model still achieves a smoother result (23.7\% reduction in acceleration error) while improving greatly in MPJPE by 25.4\%. 

\begin{table}[!thb]
\caption{\textbf{Testing error of state-of-the-art models on 3DPW, MPI-INF-3DHP, and H3.6M.} Here we compare with state-of-the-art methods on video pose estimation, and report metrics on positional accuracy (MPJPE and PA-MPJPE) as well as acceleration error. Notice that since an important annotation of H3.6M dataset has since been made unavailable, we put all methods that are trained with such supervision in italics. The most fair comparison is between our method and VIBE \cite{vibe}, the previous best-performing model (trained using the same datasets). } \label{t:motion_result} 

\centering
\resizebox{\textwidth}{!}{%
\begin{tabular}{l|c|c|c|c|c|c|c|c|c|c|c|c|c|c|r}
\hline
\multicolumn{1}{c|}{\multirow{2}{*}{}} & \multicolumn{3}{c|}{3DPW } & \multicolumn{3}{c|}{MPI-INF-3DHP } & \multicolumn{3}{c|}{H3.6M } \\ \cline{2-10} 
\multicolumn{1}{c|}{} & MPJPE $\downarrow$ & PA-MPJPE $\downarrow$ &  ACC-ERR $\downarrow$ & MPJPE $\downarrow$ & PA-MPJPE $\downarrow$ &  ACC-ERR $\downarrow$ & MPJPE $\downarrow$ & PA-MPJPE $\downarrow$ &  ACC-ERR $\downarrow$ \\ \hline
     
     \textit{HMR  (w/ H3.6M SMPL)} \cite{hmr} & \textit{130.0} & \textit{76.7}  & \textit{37.4} & \textit{124.2} & \textit{89.8} & - & \textit{88} & \textit{56.8} & -    \\
     
     \textit{HMMR (w/ H3.6M SMPL)} \cite{hmmr} & \textit{116.5} & \textit{72.6} & \textit{15.2} &  -  & - & - & - & \textit{56.9} & -    \\
     
     \textit{SPIN (w/ H3.6M SMPL)} \cite{spin}  & \textit{96.9} & \textit{59.2} & \textit{29.8} &  \textit{105.2}  & \textit{67.5} & - & - & \textit{41.1} & -    \\
     
     \textit{Sun et al. (w/ H3.6M SMPL)} \cite{1908.07172}  & - & \textit{69.5} & - &  - & - & - & \textit{59.1} & \textit{42.4}  & -    \\
     
     \textit{VIBE (w/ H3.6M SMPL)} \cite{vibe}  & \textit{82.9} & \textit{51.9} & \textit{23.4} &  \textit{96.6} &  \textit{64.6} & \textit{31.2} & \textit{65.6} & \textit{41.4}  & \textit{27.3  }  \\ \hline
     
     VIBE (w/o H3.6M SMPL) \cite{vibe}  & 91.9 & 57.6 & 25.4 &  103.9  & 68.9 & 27.3 & 78.0 & 53.3 & 27.3  \\
     \hline
     MEVA (w/o H3.6M SMPL)  & 86.9 & 54.7 & \textbf{11.6} & \textbf{96.4} & 65.4 & \textbf{11.1}  & 76.0 & 53.2 & \textbf{15.3 }   \\
     \hline
\end{tabular}}
\end{table}

\subsubsection{Qualitatively Results}

As motions are best seen in videos, please refer to \href{https://youtu.be/YBb9NDz3ngM}{the supplementary video} for qualitative results. Overall, our model achieves better acceleration error while preserving high joint position accuracy, resulting in an overall smooth and natural motion.

\subsection{Ablation Experiments}
\label{abla}

\subsubsection{Effect of data augmentation for training the motion VAE}
\label{abla_augmentation}
As mentioned in Sec.\ref{augmentation}, data augmentation performed on the AMASS dataset significantly improves the generalizability of our motion VAE. Table \ref{t:vae_abla} demonstrates the VAE reconstruction error on the unseen sequences from the 3DPW dataset (train/test/val), with various levels of data augmentation techniques. Overall, RR (random root) rotation is essential since the motion sequences in AMASS dataset, captured mostly in MoCap studio, have a single fixed initial root rotation. Model trained only on AMASS would suffer greatly when it encounters variation in root rotation. Changing the sampling frame rate (FR) and flip left and right (LR) also provide a significant boost to generalizability. Using all three augmentation techniques results in our best performing motion VAE. 

\begin{table}[!thb]
\caption{\textbf{Ablation of Data augmentation for VAE.} Here we show the effects of data augmentation on the VAE. RR: random root rotation, FR: different sampling framerate, LR: left and right flip}
\label{t:vae_abla} 
\centering
\resizebox{\textwidth}{!}{%
\begin{tabular}{l|c|c|c|c|c|c|c|c|c|c|c|c|c|c|r}
\hline
\multicolumn{1}{c|}{\multirow{2}{*}{}} & \multicolumn{3}{c|}{Train } & \multicolumn{3}{c|}{Val } & \multicolumn{3}{c|}{Test} \\ \cline{2-10} 
\multicolumn{1}{c|}{} & MPJPE $\downarrow$ & PA-MPJPE $\downarrow$ &  ACC-ERR $\downarrow$ & MPJPE $\downarrow$ & PA-MPJPE $\downarrow$ &  ACC-ERR $\downarrow$ & MPJPE $\downarrow$ & PA-MPJPE $\downarrow$ &  ACC-ERR $\downarrow$ \\ \hline
     No Aug  & 200.2 & 86.0 & 15.9 & 181.7 & 90.5 & 15.8 & 172.8 & 75.2 & 14.1 \\
     RR & 146.3 & 78.0 &  12.1 & 133.9  & 81.2 & 13.0 & 115.0 & 70.4 & 11.0 \\
     FR + LR & 291.0 & 63.03 &  14.7 & 308.8  & 66.1 & 16.1 & 261.9 & 58.9 & 13.3 \\
     \hline
     FR + LR +  RR & \textbf{72.7} & \textbf{52.3} &  \textbf{8.9} & \textbf{72.1}  & \textbf{52.9} & \textbf{9.3} & \textbf{58.7} & \textbf{43.9} & \textbf{8.2} \\
     \hline
\end{tabular}}
\end{table}

\subsubsection{Coarse motion vs fine motion retrieval}
\begin{figure}[ht]
    \centering
    \includegraphics[width=\linewidth]{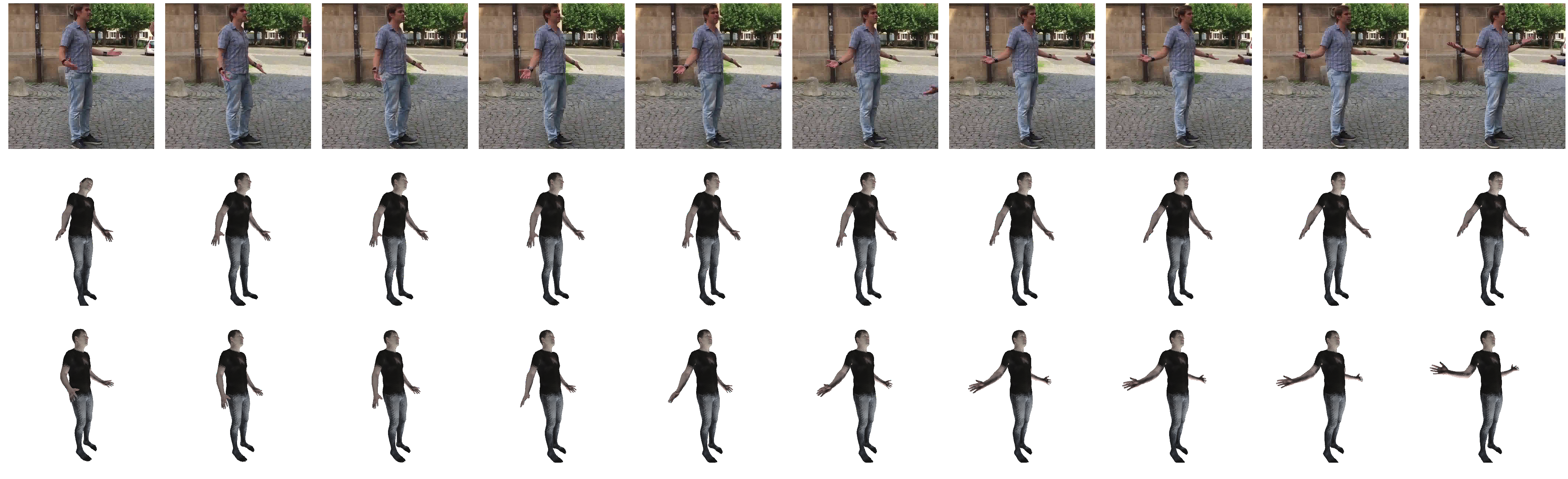}
    \caption{\textbf{Breakdown of coarse and fine human motion.} The first row of estimated human is the coarse part of the motion (output of VME), while the second row adds back the fine details (output of MRR).}
    \label{fig:highlow}
\end{figure}

\textit{MEVA} benefits from an explicit breakdown of coarse and fine motion retrieval, using a temporal compressive step that captures the overall motion in a given human motion sequence. Just how much coarse/smooth motion information is retrieved in the framework?
Table \ref{t:motion_hl_abla} shows the result of \textit{MEVA} if trained only using the VME (capturing only coarse motion). Notice that \textit{MEVA} with only VME achieves result similar to HMMR \cite{hmmr}, the previous state-of-the-art in producing low acceleration error motion estimation. An illustrative visualization of coarse and fine motion decomposition can be found in Fig.\ref{fig:highlow}.

\begin{table}[!thb]
\caption{\textbf{Ablation of MEVA.} Here we show the \textit{MEVA} model trained with only Variational Motion Estimator (without Motion Residual Regressor) or without using pretrained VAE. } \label{t:motion_hl_abla} 
\centering
\resizebox{3 in}{!} {
\begin{tabular}{l|c|c|c|c|c|r}
\hline
\multicolumn{1}{c|}{\multirow{2}{*}{}} & \multicolumn{3}{c|}{3DPW } \\ \cline{2-4} 
\multicolumn{1}{c|}{} & MPJPE $\downarrow$ & PA-MPJPE $\downarrow$ &  ACC-ERR $\downarrow$ \\ \hline
    MEVA-VME only   & 118.1 & 73.7 & 15.4 \\ \hline
    MEVA-without using pretrained VAE   & 95.9 & 59.7 & 14.1 \\ \hline
     MEVA   & \textbf{86.9} & \textbf{54.7} & \textbf{11.6}  \\
     \hline
\end{tabular}}
\end{table}

\subsubsection{Effects of the pretrained Motion VAE}
\label{abla_pretrained_vae}
\begin{figure}[t]
    \centering
    \includegraphics[width=0.85\linewidth]{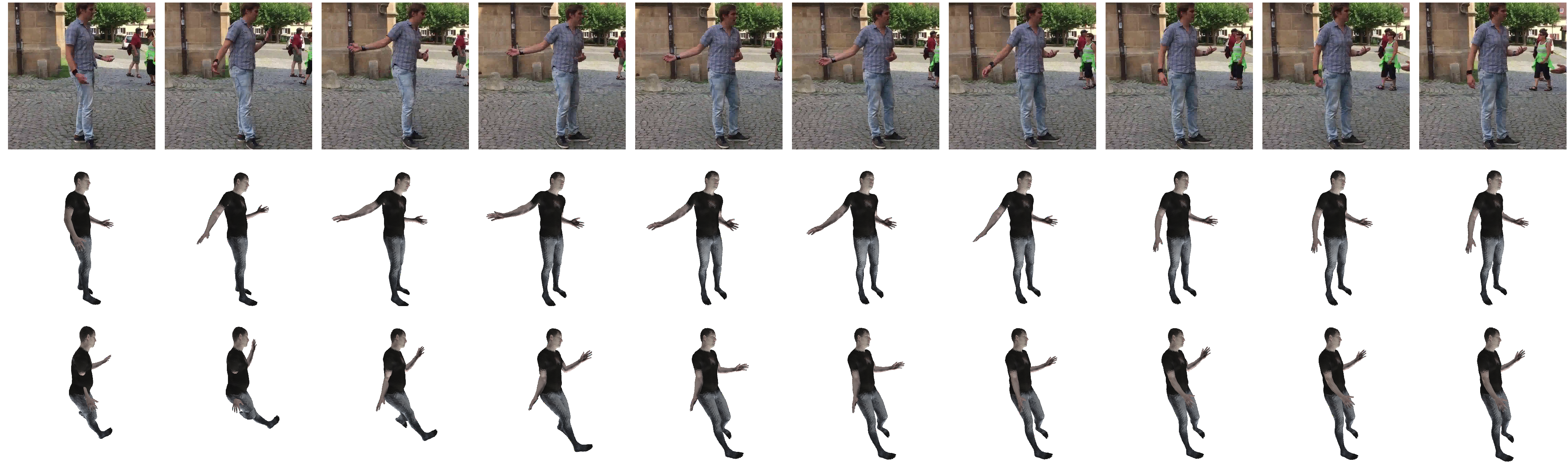}
    \caption{\textbf{\textbf{MEVA} results without using pretrained VAE.} Here we show an ablative study where we do not use a pretrained VAE. The second row shows \textbf{MEVA} result without using a pretrained VAE and the first row shows our full model estimation. Notice that the model estimates unnatural human poses in the first few frames. }
    \label{fig:abla}
\end{figure}

\textit{MEVA} benefits from using a pretrained motion VAE's latent space. As argued in Sec.\ref{compact}, using a pretrained VAE provides a human motion subspace that assists in constraining the estimated motion sequences to be natural and plausible human motion. Table \ref{t:motion_hl_abla} shows the result of \textit{MEVA} trained without using a pretrained VAE (not using the pretrained $D_{vae}$). In this case, the whole framework is trained end-to-end from scratch. Here we observe that the model performed relatively well in both accuracy and smoothness, demonstrating the power of our two-stage estimation framework. However, upon visual inspection, as shown in Fig.\ref{fig:abla}, a few kinematically invalid human poses are estimated during the sequence, resulting in an overall accurate but flawed estimation.

\section{Conclusion}
We have shown that to achieve temporally smooth and accurate 3D human pose estimates, it is important to learn a compressive model that encodes the smoothness of general human motion while also learning an image-based regression model that can capture person-specific motion. We propose a two-stage model that first trains a Variational Autoencoder to model the general statistics of coarse/smooth human motion and then learns a person-specific motion refinement regression module to retain motions not captured by the general motion model. Through comprehensive experiments, we demonstrate that our method produces both smooth and accurate motion. \\

\textbf{Acknowledgements:} This project was sponsored in part by IARPA \\ (D17PC00340), and JST AIP Acceleration Research Grant (JPMJCR20U1).

\bibliographystyle{splncs}
\bibliography{main}





\title{3D Human Motion Estimation via \\Motion Compression and Refinement\\ Supplementary Material} 
\titlerunning{3D Human Motion Estimation}
%
\author{Zhengyi Luo \and S. Alireza Golestaneh \and Kris M. Kitani}
\authorrunning{Z. Luo et al.}

\institute{Carnegie Mellon University, Pittsburgh, PA 15213, USA  \\
\email{\{zluo2,sgolesta,kkitani\}@cs.cmu.edu}}

\maketitle
This supplementary material is organized as follows.  
In Section 1, we provide qualitative results for our proposed method (\textit{MEVA}).  In Section 2, we provide complementary ablation studies. In Section 3, we will discuss the failure modes of our method. In the last section, we include the details of our implementation.

\section{Qualitative Results}
To best view our motion estimation and compare it with state-of-the-art, please refer to the \href{https://youtu.be/YBb9NDz3ngM}{supplementary video}.

Specifically, in the supplementary video, we first show a visual demonstration of our two-stage decomposition of coarse and fine motion from a given video sequence. Next, we demonstrate the qualitative comparison between our algorithm and the prior state-of-the-art (VIBE\cite{vibe}) and show that our method achieves smoother, more natural, and accurate motion estimation.
Finally, we will discuss the implementation details of our method.

\section{Additional Ablation Studies}

\subsection{Comparison with Average Filtering}
 While our method has significantly reduced the acceleration error and achieves state-of-the-art accuracy, one can still apply postprocessing to existing sequences to further improve the prediction. To best study its effects, here we implement a simple average filter using spherical linear interpolation (slerp) in quaternion. Specifically, for each joint rotation in SMPL $q^i$ at timestep $t$, we apply slerp with a ratio of 0.5:  $q^i_t = slerp(q^i_t, q^i_{t+1}, 0.5) $. Table \ref{t:smooth_abla} shows the result of applying averaging filtering as postprocessing on both VIBE \cite{vibe} and \textit{MEVA}. From the results, it is clear that average filtering can help reduce the acceleration error of both VIBE and \textit{MEVA} while slightly affecting accuracy. It is also conceivable that more sophisticated methods such as solving a constrained optimization problem \cite{Huang,Peng2018,Mehta2017} can further improve results. Nonetheless, in the paper we only compare with feed-forward methods without any postprocessing, since postprocessing approaches are complementary to feed-forward methods and would be beneficial to all of them.

\begin{table}[!thb]
\caption{\textbf{Ablation study on average filtering.} Here we show the result of applying the average filter on the output from VIBE \cite{vibe} and \textit{MEVA}. } \label{t:smooth_abla} 
\centering
\resizebox{3 in}{!} {
\begin{tabular}{l|c|c|c|c|c|r}
\hline
\multicolumn{1}{c|}{\multirow{2}{*}{}} & \multicolumn{3}{c|}{3DPW } \\ \cline{2-4} 
\multicolumn{1}{c|}{} & MPJPE $\downarrow$ & PA-MPJPE $\downarrow$ &  ACC-ERR $\downarrow$ \\ \hline
     VIBE (w/o H3.6M SMPL) \cite{vibe}  & 91.9 & 57.6 & 25.4  \\
     VIBE (w/o H3.6M SMPL)  \cite{vibe} + Average Filtering & 91.6 & 57.8 & 13.5  \\
     MEVA (w/o H3.6M SMPL)  (ours) & \textbf{86.9} & \textbf{54.7} & 11.6  \\
     MEVA (w/o H3.6M SMPL) + Average Filtering (ours) & 87.6 & 55.5 & \textbf{8.2}  \\
     \hline
\end{tabular}}
\end{table}

\subsection{Effects of a long temporal window}

\textit{MEVA} uses a significantly longer temporal window (90 frames) than prior art (HMMR \cite{hmmr}: 20 frames, VIBE\cite{vibe}: 16 frames). To show that our \textit{MEVA} framework benefits more from this setting, we retrain VIBE with a 90 frames temporal window. As shown in Table \ref{t:tp_abla}, using the same size temporal window, \textit{MEVA} produces better results on all three metrics and maintains a significant advantage in acceleration error. Notice that VIBE trained with a longer temporal window shows a slight improvement against the ones that use a shorter window, validating our intuition that a longer temporal window provides a more substantial context for motion estimation. Nonetheless, our two-stage decomposition method is more effective in utilizing a longer temporal window due to its separate motion compression and refinement stages.

\begin{table}[!thb]
\caption{\textbf{Ablation study on temporal window size.} Here we show the results of using different temporal windows in VIBE \cite{vibe} and \textit{MEVA} } \label{t:tp_abla} 
\centering
\resizebox{3 in}{!} {
\begin{tabular}{l|c|c|c|c|c|r}
\hline
\multicolumn{1}{c|}{\multirow{2}{*}{}} & \multicolumn{3}{c|}{3DPW } \\ \cline{2-4} 
\multicolumn{1}{c|}{} & MPJPE $\downarrow$ & PA-MPJPE $\downarrow$ &  ACC-ERR $\downarrow$ \\ \hline
     VIBE (w/o H3.6M SMPL) + 16 frames \cite{vibe}  & 91.9 & 57.6 & 25.4  \\
     VIBE (w/o H3.6M SMPL) + 90 frames \cite{vibe}  & 88.1 & 56.6 & 21.2  \\
     MEVA (w/o H3.6M SMPL) + 90 frames (ours) & \textbf{86.9} & \textbf{54.7} & \textbf{11.6}  \\
     \hline
\end{tabular}}
\end{table}

\subsection{Effects of STE and VME on \textit{MEVA}}
Here we take a further look into the effects of different components (STE, VME, and MRR) of our proposed method. Notice that without the Variational Motion Estimator (VME), our method will collapse into a single-stage estimator that only relies on the SMPL regressor, which has been studied extensively in prior art. Thus, here we only study the effects of Spatial Temporal Feature Extractor (STE) and Motion Refinement Regressor (MRR). Table \ref{t:meva_abla_ste_mrr} shows the results of our framework trained without STE or MRR. Without the STE, \textit{MEVA} obtains high accuracy but suffers from high acceleration error. This indicates that STE produces correlated features that impart the necessary temporal consistency information to MRR. We reason that without STE, even although initialized with coarse estimation from VME, MRR will be biased by the input visual features and produce a temporally inconsistent refinement pose that negatively affects the overall estimation. On the other hand, without MRR, our method is reduced to one stage and only estimates the coarse motion.  As shown in the result, using only VME will lead to an overly smoothed motion estimation and result in a higher acceleration error (underestimating movement also leads to high acceleration error).

\begin{table}[!thb]
\caption{\textbf{Ablation of MEVA components.} Here we show \textit{MEVA} trained without STE (with both VME and MRR) and without MRR (with both STE and VME).} \label{t:meva_abla_ste_mrr} 
\centering
\resizebox{3 in}{!} {
\begin{tabular}{l|c|c|c|c|c|r}
\hline
\multicolumn{1}{c|}{\multirow{2}{*}{}} & \multicolumn{3}{c|}{3DPW } \\ \cline{2-4} 
\multicolumn{1}{c|}{} & MPJPE $\downarrow$ & PA-MPJPE $\downarrow$ &  ACC-ERR $\downarrow$ \\ \hline
    MEVA w/o STE   & 89.7 & 55.4 & 29.0 \\ 
    MEVA w/o MRR   & 118.1 & 73.7 & 15.4 \\ 
     MEVA   & \textbf{86.9} & \textbf{54.7} & \textbf{11.6}  \\
     \hline
\end{tabular}}
\end{table}

\section{Failure Modes}
Although \textit{MEVA} shows promising results in producing smooth and accurate human motion, there is still room for improvement. 

\subsection{Sliding window processing}
\begin{figure}[ht]
    \centering
    \includegraphics[width=0.85\linewidth]{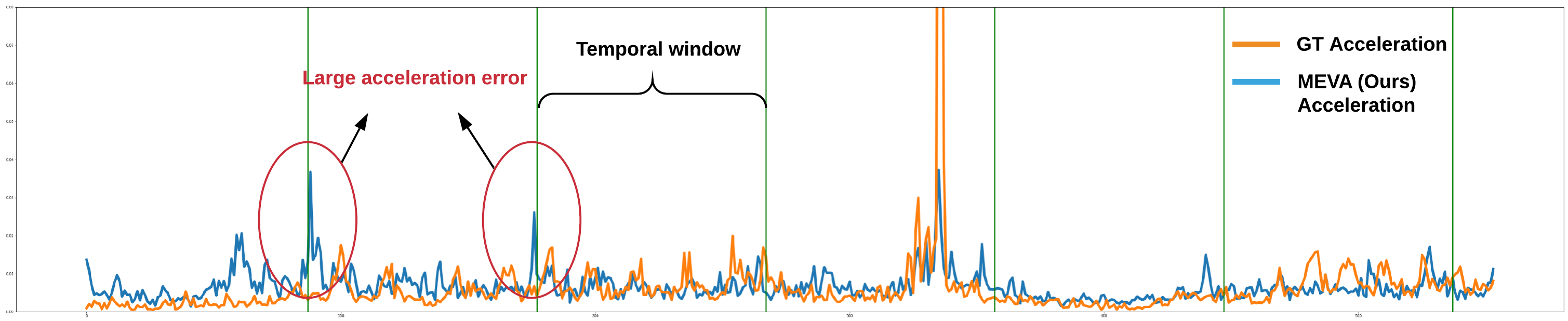}
    \caption{\textbf{Sliding window failure mode.} This plot shows that at the intersection of temporal windows, \textit{MEVA} can result in a inaccurate transition and bring a large acceleration error. Each green line in the plot marks a temporal window, and there are large spikes of acceleration error at the first two intersections.}
    \label{fail_window}
\end{figure}

\textit{MEVA} processes videos using a sliding window: input video sequences are splited into chunks of 90 frames for processing. Due to the natural of this sliding window approach, inconsistency can sometimes be observed at a 3 second interval (videos are assumed to be at 30 fps). The explanation is as follows: the coarse motion estimated by VME can be quite different between each temporal window and MRR sometimes is unable to make enough adjustments to account for a smooth transition. Each temporal window also has their own STE, so the features from each window are longer correlated. Fig. \ref{fail_window} shows an instance of this behavior. For visual inspection, please refer to our supplement video. 

\subsection{Occluded body parts}
Occluded body parts can still be challenging for \textit{MEVA}. During occlusion, the lack of visual indicators will compel \textit{MEVA} to rely on coarse motion estimation over the whole sequence and leads to a misscapture of detailed motion. Please refer to the supplementary video for an example.

\subsection{Missing hands and face movement}
Since the original SMPL\cite{SMPL} model does not contain joints for the hand and face, all methods using SMPL do not capture hand movements and facial expressions. Moreover, there is not enough high quality 3D data that provides hand and face annotations. A recent work \cite{smplx} develops an enhanced SMPL model that jointly models body pose, hands, and face, but this model has not gained significant traction in the pose estimation community. We believe that capturing hands and face movement in motion estimation is an essential direction for future work.
\vspace{-0.5mm}
\section{Implementation Details}
\vspace{-0.5mm}
\subsection{Human Motion VAE}
The motion VAE's encoder, $E_{vae}$, is a bidirectional Gated Recurrent Unit (bi-GRU) with average pooling to obtain the temporal encoding $h$ of the overall input motion sequence $M_W \in R^{W\times 144}$. We pass the temporal encoding $h$ into a multilayer perceptron (MLP) with two hidden layers (1024, 512) and two heads to obtain the mean $\mu$ and variance $\sigma$ for the latent code $z$. For the decoder $D_{vae}$, a forward GRU is used to decode the output motion sequence. At each time step, the GRU takes in the previous step estimation $\theta_{t-1}$ and the current latent code $z \in R^{1\times S_z}$ to output a 512 latent feature. The feature is then passed through an MLP with two hidden layers (1024, 512) to generate the reconstructed pose $\theta \in R^{1\times 144}$
\vspace{-0.5mm}
\subsection{ Spatio-Temporal Feature Extractor}
\vspace{-0.5mm}
For a video input, we first preprocess the video frames using a pretrained ResNet-59 network \cite{hmr}. The feature extractor outputs $f_{i} \in R^{2048}$ for each frame. The extracted features within the same temporal window $W$ (we choose $W = 90$) are stacked together $[f_{t}]^{90}_{t = 1} \in R^{90 \times 2048}$ and are encoded by STE into a sequence of temporally correlated features $[f'_{t}]^{90}_{t = 1} \in R^{90 \times 2048}$. STE is a 2 layer bi-GRU with hidden size 1024 that outputs a feature encoding at each timestep. $E_{motion}$ that shares the same architecture as $E_{feat}$ except for the final average pooling step to come up with a latent code $z \in R^{1 \times 512}$ that represents the whole motion sequence. 

\subsection{Motion Residual Regressor}
The MRR consists of 2 fully connected layers, each with 1024 neurons. It takes in per-frame features and a set of initializing parameters (pose, shape, and camera) and iterative refines its predictions (pose, shape, and camera) for $k$ iterations.







\end{document}